\documentclass[letterpaper, 10 pt, conference]{ieeeconf}  

\IEEEoverridecommandlockouts                              

\usepackage{amsmath} 
\usepackage{amssymb}  

\usepackage{subcaption}

\usepackage{moreverb,url}
\usepackage{forest}
\usepackage{listings}
\usepackage{graphicx}

\definecolor{codegreen}{rgb}{0,0.6,0}
\definecolor{codegray}{rgb}{0.5,0.5,0.5}
\definecolor{codepurple}{rgb}{0.58,0,0.82}
\definecolor{codered}{rgb}{0.5,0,0}

\lstdefinestyle{mystyle}{
    commentstyle=\color{codered},
    keywordstyle=\color{magenta},
    numberstyle=\tiny\color{codegray},
    stringstyle=\color{codegreen},
    breakatwhitespace=false,         
    breaklines=true,                 
    captionpos=b,                    
    keepspaces=true,                 
    numbers=left,                    
    numbersep=5pt,                  
    showspaces=false,                
    showstringspaces=false,
    showtabs=false,                  
    tabsize=2,
    basicstyle=\ttfamily\footnotesize
}

\def\numtasks{100}

\newcommand{\expertpolicy}{\ensuremath{\pi^*}}
\newcommand{\action}{\ensuremath{\boldsymbol{a}}}

\newcommand{\obs}{\ensuremath{\boldsymbol{o}}}
\newcommand{\task}{\ensuremath{\mathfrak{T}}}
\newcommand{\variation}{\ensuremath{\nu}}
\newcommand{\episode}{\ensuremath{\tau}}
 
\definecolor{folderbg}{RGB}{124,166,198}
\definecolor{folderborder}{RGB}{110,144,169}

\def\Size{4pt}
\tikzset{
  folder/.pic={
    \filldraw[draw=folderborder,top color=folderbg!50,bottom color=folderbg]
      (-1.05*\Size,0.2\Size+5pt) rectangle ++(.75*\Size,-0.2\Size-5pt);  
    \filldraw[draw=folderborder,top color=folderbg!50,bottom color=folderbg]
      (-1.15*\Size,-\Size) rectangle (1.15*\Size,\Size);
  }
}

\title{\LARGE \bf
RLBench: The Robot Learning Benchmark \& Learning Environment
}

\author{Stephen James$^{1}$, Zicong Ma$^{2}$, David Rovick Arrojo$^{2}$, Andrew J. Davison$^{1}$
\thanks{$^{1}$Dyson Robotics Lab, Imperial College London}%
\thanks{$^{2}$UROP, Imperial College London}%
}

\begin{document}

\maketitle
\thispagestyle{empty}
\pagestyle{empty}

\begin{abstract}
We present a challenging new benchmark and learning-environment for robot learning: RLBench. The benchmark features \numtasks{} completely unique, hand-designed tasks ranging in difficulty, from simple target reaching and door opening, to longer multi-stage tasks, such as opening an oven and placing a tray in it. We provide an array of both proprioceptive observations and visual observations, which include rgb, depth, and segmentation masks from an over-the-shoulder stereo camera and an eye-in-hand monocular camera. Uniquely, each task comes with an infinite supply of demos through the use of motion planners operating on a series of waypoints given during task creation time; enabling an exciting flurry of demonstration-based learning. RLBench has been designed with scalability in mind; new tasks, along with their motion-planned demos, can be easily created and then verified by a series of tools, allowing users to submit their own tasks to the RLBench task repository. This large-scale benchmark aims to accelerate progress in a number of vision-guided manipulation research areas, including: reinforcement learning, imitation learning, multi-task learning, geometric computer vision, and in particular, few-shot learning. With the benchmark's breadth of tasks and demonstrations, we propose the first large-scale few-shot challenge in robotics. We hope that the scale and diversity of RLBench offers unparalleled research opportunities in the robot learning community and beyond. Benchmarking code and videos can be found here\footnote{\url{https://sites.google.com/view/rlbench}}.
\end{abstract}

\section{INTRODUCTION}

Robot manipulation systems broadly fall somewhere on a spectrum ranging from traditional, modular methods, that include object recognition, state estimation, and planning, to fully end-to-end approaches that leverage deep learning and large-scale data to learn a mapping from input observations directly to motor actions, with the intuition that the `traditional' modules are embedded in the weights of a deep neural network. Driven by the successful combination of large-scale data \cite{ILSVRC15} and deep learning algorithms in the field of computer vision \cite{krizhevsky2012imagenet}, there is now a large body of work looking at increasing the capabilities of robotic agents through the use of reinforcement learning  \cite{james20163d, kalashnikov2018qt}, meta-learning \cite{finn2017one, james2018task, yu2018one}, multi-task learning \cite{devin2017learning, hausman2018learning}, etc. However, there is currently no standard in place for comparing manipulation methods in these respective areas. Although there exist benchmarks such as OpenAI Gym \cite{brockman2016openai} and DeepMind Control Suite \cite{tassa2018deepmind} for evaluating continuous-control reinforcement learning algorithms, their focus is not on real-world problems, and it is often the case that algorithms in these toy-benchmarks do not scale to more complex, real-world tasks. Few-shot learning methods for robotics also suffer from a lack of well defined tasks; for example, in Finn \textit{et al.} \cite{finn2017one} and James \textit{et al.} \cite{james2018task} there is a very narrow distribution of tasks, where the task of ``placing a peach into a red bowl'' would be considered a different task to ``placing an apple in to a green bowl''. Despite the increase in these data-driven approaches, it is not clear where the ideal location on this `learning' spectrum lies for complex robotics tasks that we may one day want robots performing in our homes. Given all of these problems, there seems to be a need for a benchmark that evaluates not only the diverse range of robot learning fields that are now emerging, but also a range of visually-guided manipulation approaches from both sides of the spectrum.

\begin{figure*}
  \centering
  \includegraphics[width=0.99\linewidth]{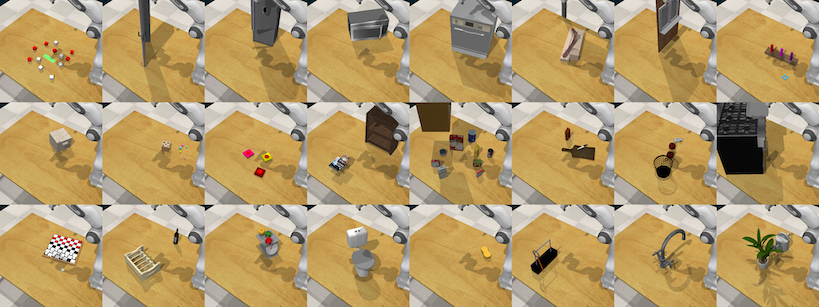}        
  \caption{RLBench is a large-scale benchmark consisting of \numtasks{} completely unique, hand-designed tasks. In this figure we show a sample of 24 tasks that feature in the benchmark. Example tasks include stacking a set of 6 colored blocks in a pyramid (top left), inserting a shape onto a peg (top right), finish setting up a checkers board (bottom left), and watering a plant (bottom right). To get a better understanding of the variety of tasks, please watch the video.}
  \label{fig:task_grid}
\end{figure*}

This motivates the need for a one-size-fits-all benchmark that allows the capability to utilise large-scale data, whilst also allowing classical systems to be compared. To that end, we present RLBench, which is an ambitious large-scale benchmark and learning environment designed to facilitate research in a number of both classical and deep-learning based robot manipulation areas. RLBench is deliberately highly challenging and forward looking.
The benchmark includes \numtasks{} completely unique, hand-designed tasks ranging in difficulty (shown in Figure \ref{fig:task_grid}), which share a common Franka Emika Panda robot arm, featuring a range of sensor modalities, including joint angles, velocities and forces, an eye-in-hand camera and an over-the-shoulder stereo camera setup. Each of the \numtasks{} tasks comes with a number of textual descriptions and an infinite set of demonstrations made possible through our task building tools that use waypoint-based motion planning.

In this paper, we discuss a host of research areas that would benefit from this benchmark, including, but not restricted to, reinforcement learning, imitation learning, few-shot learning, multi-task learning, and geometric based methods, such as SLAM. In addition to the benchmark, we also contribute an open-source set of tools that will allow rapid development of new tasks (through the use of PyRep \cite{james2019pyrep}) in order to improve the size and scope of the benchmark over time. To summarise, RLBench has the following 3 key aims: 

\begin{itemize}
    \item Provide a benchmark and learning environment for both `robot learning' and `traditional' methods.
    \item Provide the a large-scale few-shot challenge, where given \textit{M} training tasks and \textit{N} unseen tasks, a system must take \textit{K} different demonstrations of each of the \textit{N} unseen tasks, and then be able to perform these tasks in new configurations. 
    \item Provide a set of tools to allow easy task creation. 
\end{itemize}

\section{Related Work}

We review existing datasets, benchmarks, and learning environments that could be considered similar to ours in an effort to further motivate RLBench. Firstly we cover reinforcement learning benchmarks, followed by benchmarks designed specifically for manipulation.

\paragraph{\textbf{Reinforcement Learning}}

Largely as a consequence of the seminal work that saw an algorithm learn to play a range of Atari 2600 video games to superhuman level directly from image pixels \cite{mnih2015human}, deep reinforcement learning (DRL) has increasingly become prevalent in the literature, leading to a number of recent further success in the games of Go \cite{silver2016mastering}, Chess \cite{silver2017mastering}, StarCraft \cite{alphastar}, and Dota \cite{OpenAI_dota}. With the success of these approaches, there has been a surge in developing DRL algorithms to solve continuous control environments \cite{lillicrap2015continuous, schulman2015trust, schulman2017proximal, haarnoja2018soft, fujimoto2018addressing}. These learned (continuous control) agents are usually tested on benchmarks such as OpenAI Gym \cite{brockman2016openai} or the DeepMind Control Suite \cite{tassa2018deepmind}. 
However, apart from a small number of robotic tasks in OpenAI Gym, these benchmarks feature only toy tasks that often do not resemble real-world problems that robots will need to overcome. To combat this, many projects create their own manipulation tasks to evaluate their approach, making comparisons difficult. As a direct consequence of this, these created tasks can often succumb to unintentionally introducing another hyperparameter into the method in the form of the task design itself. For example, a method could fail on a more challenging task, and so results would only be presented for a simpler set of tasks. This is something a standard benchmark of tasks could alleviate.
(We should mention the very recently announced
Meta-World project \cite{Meta-World}, a multi-task benchmark for meta-learning research in manipulation, though full documentation describing the aims of that project is not available at the time of writing.)

\paragraph{\textbf{Manipulation}}

Most related work in benchmarking robot manipulation algorithms often concentrates on solving only one of the manipulation sub-problems, focusing on either perception, grasping, or planning. But first, we look at benchmarks that evaluate the system as a whole. The Amazon Robotics Challenge (ARC) \cite{eppner2016lessons} was an attempt to create a benchmark for robotic picking and stowing. Although it was a successful challenge that drew many conclusions, such as the usefulness of a dual gripper and suction cup end-effector \cite{morrison2018cartman}, it was difficult to reproduce in a lab setup. The ACRV Picking Benchmark \cite{leitner2017acrv} aimed to solve this by creating a similar, but reproducible setup to the ARC. The issue with picking and stowing is that it is but one of many possible tasks; RLBench on the other hand comes with \numtasks{} unique tasks, many of which involve some aspect of picking and placing. Similarly to ARC, the RoboCup@Home competition \cite{wisspeintner2009robocup} is run annually, but has a greater range of tasks that must be completed. However, given that no large-scale data is given beforehand, this makes reinforcement learning and other end-to-end approaches difficult to apply in the competition. RLBench is a platform that can unify both old and new methods and compare them on an even playing field.

For evaluating imitation learning systems in particular, RoboTurk \cite{mandlekar2018roboturk} was a recent attempt to leverage crowd sourcing to obtain data for tasks, but because of this the system has only three tasks. Whilst RoboTurk is entirely in simulation, Simitate \cite{memmesheimer2019simitate} on the other hand is a hybrid approach, where real world observations (RGB-D camera calibrated against a motion capturing system) are combined with a simulated environment for benchmarking. In contrast to RoboTurk, we do not crowd source our demonstrations, but instead rely on an infinite supply of generated demonstrations collected via motion planners. Although Simitate offers the benefit of being partially a real-world dataset, the addition of new tasks requires time-consuming calibration and motion capturing; our system on the other hand sacrifices the real-world aspect, but in exchange we receive the ability generate a diverse range of tasks in a scalable way.

Moving on from whole-system benchmarks, there are a host of benchmarks that focus on sub-problems, for example perception datasets, from both the computer vision community (such as ILSVRC \cite{ILSVRC15}, COCO \cite{lin2014microsoft}, Pascal-VOC \cite{everingham2015pascal}, etc), and the robotics community (such as BigBIRD \cite{singh2014bigbird}, YCB-Video \cite{xiang2017posecnn}, etc). For grasping, both OpenGrasp \cite{ulbrich2011opengrasp} and VisGraB \cite{popovic2011grasping} are popular simulation-based benchmarks, whilst the YCB dataset \cite{calli2015benchmarking} focuses on real-world objects. In comparison to these, RLBench allows robotic systems to be evaluated on the complete robotic pipeline, rather than limited to sub-problems such as object detection, state estimation, grasp selection, and planning.

\section{Benchmark Properties}

\begin{figure}
  \centering
  \includegraphics[width=0.9\linewidth]{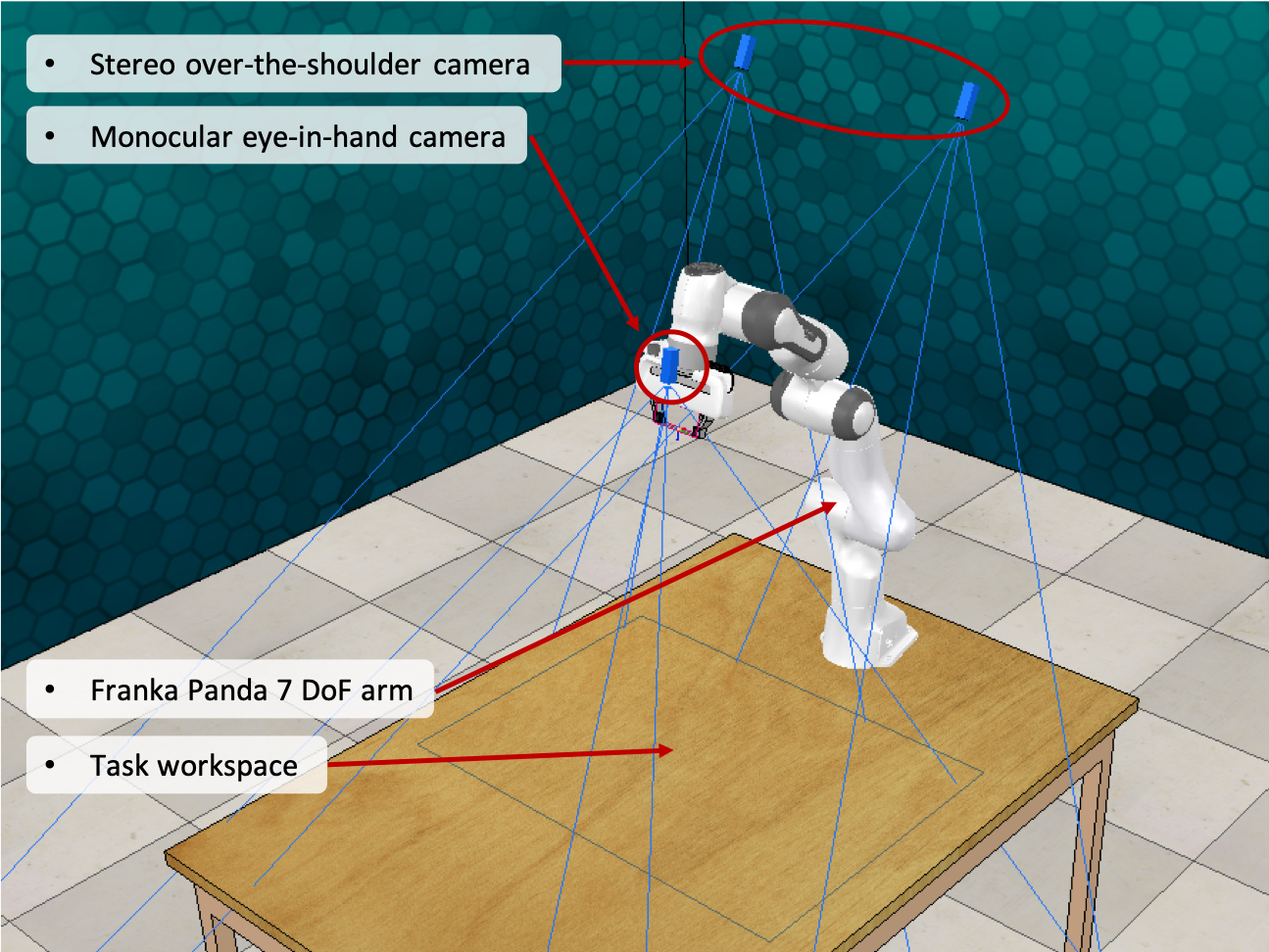}        
  \caption{The V-REP scene consists of a Franka Panda affixed to a wooden table, surrounded by 3 directional lights. Observations include rgb, depth, and segmentation masks from an over-the-shoulder stereo camera and a eye-in-hand monocular camera, along with robot proprioceptive data, which includes joint angles, velocities, and torques, and the gripper pose. The arm can be easily swapped out for another arm if required.}
  \label{fig:panda_scene_label}
\end{figure}

\begin{figure}
  \centering
  \includegraphics[width=1.0\linewidth]{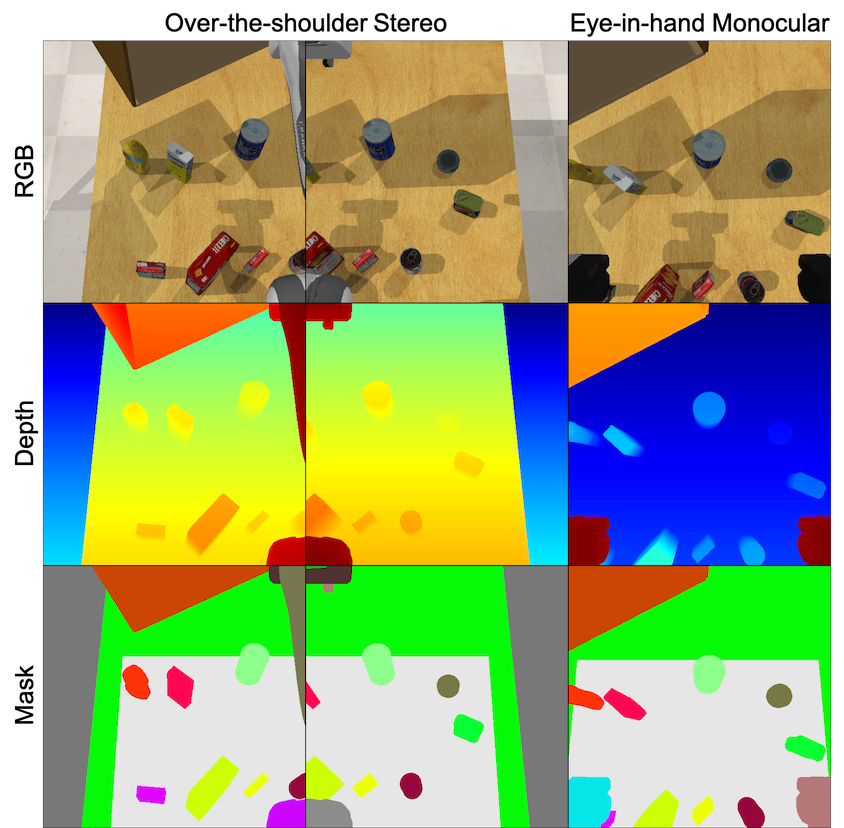}        
  \caption{A sample of the visual observations given from both the over-the-shoulder stereo and eye-in-hand monocular cameras, which supply rgb, depth, and mask images.}
  \label{fig:observations}
\end{figure}

\begin{figure*}
  \centering
  \includegraphics[width=0.99\linewidth]{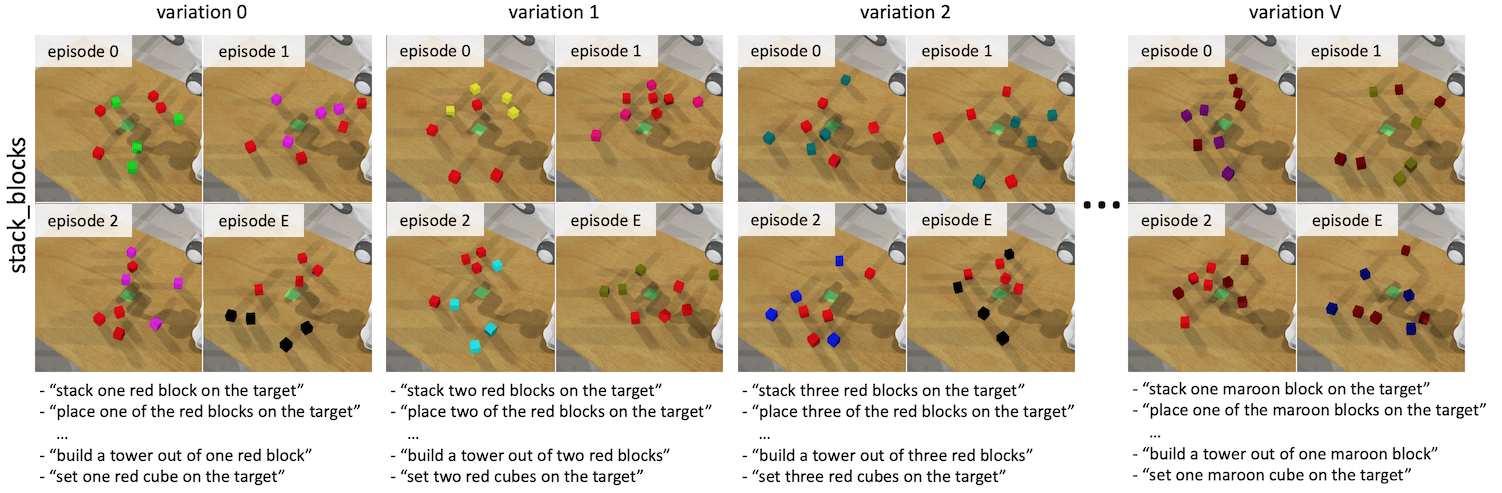}        
  \caption{An example showing the distinction between task, variation, and episode. In this case, the `\textit{stack\_blocks}' task has $V$ variations, each with $E$ episodes. Each variation comes with a list of textual descriptions that describes the objective. Across variations, usually target objects or colours are changed, whereas across episodes positions are changed.}
  \label{fig:task_ep_var_example}
\end{figure*}

When designing RLBench, we have prioritised several key properties:

\paragraph{\textbf{Diversity}} 

Algorithms we develop should be general. In order to effectively learn inter-task relationships, a truly diverse range of tasks is needed to help avoid over-fitting. 

\paragraph{\textbf{Reproducibility}} 

Reproducibility is challenging in robotics as each lab has their own robotic setup. Moving to simulation solves this, but at the risk of developing solutions that may not run as well in the real-world. However, with the rise of deep-learning methods becoming more prominent in robotics, we believe it is important to find the potential and limits of these methods in a controlled, reproducible environment. 

\paragraph{\textbf{Scale}} 

Given the amount of data modern machine learning methods need, it is important to not only have a large collection of tasks, but also the ability to produce a large number of demonstrations from these tasks. 

\paragraph{\textbf{Extensibility}} 

Following on from the previous point, we hope to continue to grow this repository of tasks. Therefore it is crucial that the task building system is as easy as possible to use. By leveraging the recently released robotics toolkit, PyRep \cite{james2019pyrep}, we are able to make a broad range of tasks in a short amount of time.

\paragraph{\textbf{Tiered Difficulty}} 

Attempting to get robots to do a single task can be challenging let alone expecting them to do numerous tasks. We therefore wanted to have a range of tasks, including both easy tasks, such as reaching, which would be well suited to new and emerging methods, to more challenging, long-time-horizon tasks that can stress-test well known state-of-the-art algorithms in use today.

\paragraph{\textbf{Realism}} 

Although we cannot claim full photorealism in our rendering system, or general realistic physics, we have put substantial effort into high quality components such as using a realistic robot model, graphics with lighting and shadows and a domain randomisation rendering option in order to maximise the potential for research on sim-to-real transfer.

\section{RLBench}

RLBench is an ambitious project which we hope to grow over many years. The benchmark and learning environment is built around a V-REP \cite{rohmer2013v} and PyRep \cite{james2019pyrep} interface. PyRep is a toolkit for robot learning research, built on top of V-REP that features a number of improvements, including speed, rendering, and flexible a API for robot control and scene manipulation. Using the combination of these two libraries, we have been able to build this ambitious benchmark, which we now describe in greater detail.

\subsection{Scene}

The V-REP scene, shown in Figure \ref{fig:panda_scene_label}, remains constant across all tasks and contains the Franka Emika Panda 7 DoF arm affixed to a wooden table, surrounded by 3 directional lights. As shown in Figure \ref{fig:observations}, visual observations can be perceived from a stereo camera, and a monocular wrist camera, which supply rgb, depth, and segmentation mask data on each frame. In addition to visual observations, robot proprioceptive data can be retrieved, which includes joint angles, velocities, and torques, along with the end-effector pose. Tasks are loaded into the scene and placed at the centre of the workspace. Every task starts with the same assumption that no objects are held, therefore, unlike many works in the literature, tasks that involve tools will first need to grasp the object appropriately in order to accomplish the task. Although this makes the environments considerably harder to complete, we believe it is an important assumption to make given that household robots will one day work under such conditions.

\subsection{Tasks, Variations \& Episodes}

RLBench employs 3 keys terms: \textit{Task}, \textit{Variation}, and \textit{Episode}. Each task consists of one or more variations, and from each variation, an infinite number of episodes can be drawn. Each variation of a task comes with a list of textual descriptions that verbally summarise this variation of the task, which could prove useful for human robot interaction (HRI) and natural language processing (NLP) research. A summary of this can be seen in Figure \ref{fig:task_ep_var_example}. Formally, we define an episode trajectory $\episode$ to consist of a series of observations $\obs$ and actions $\action$: $\episode = [(\obs_1, \action_1), \ldots, (\obs_T, \action_T)]$. These episodes are sampled from a variation $\episode \sim \variation$. Finally, we define each task to be a set of variations, $\task = \{ \variation_1, \cdots, \variation_N \}$.

We now motivate the need for the concept of a `variation' with an example. It is naturally difficult to come up with a precise way to differentiate between tasks given their subjective nature. For example, one could argue that ``pick up the apple'' and ``pick up the banana'' are different tasks, whilst one could also equally argue that they are the same ``pick up the X'' task. We therefore introduce the variation concept, which allows cases like the above to be grouped as very similar tasks. Moreover, given the way the task building tools are designed (discussed in Section \ref{sec:task_builder}), the variation concept allows a convenient way of getting as much from a task definition as possible, given that there is usually only a small amount of additional work needed to generate a large number of variations for a given task.

\subsection{Environment}

Users will interface with the benchmark and learning environment through the \textit{Environment} class. The Environment is the entry point and can spawn child environments, called \textit{TaskEnvironment}, for the tasks you are interested in solving. The environment API, which Figure \ref{alg:api_example} demonstrates, is modelled after a typical agent-environment reinforcement learning setup. Each task has a completely sparse reward of $+1$ which is given only on task completion. Users have a wide variety of action spaces at their disposal, which include absolute or delta joint velocities, absolute or delta joint positions, absolute or delta joint torque, absolute or delta end-effector velocities, and finally absolute or delta end-effector poses.

\subsection{Demonstrations}

RLBench, through the task building tool mentioned in Section \ref{sec:task_builder}, provides expert algorithm $\expertpolicy$ for each different task and their corresponding variations, allowing for demonstration episodes to be generated  The episodes produced via $\expertpolicy$ come from using the Open Motion Planning Library~\cite{sucan2012open}.

\subsection{Task Builder}
\label{sec:task_builder}

Two common simulation environments in the literature today are Bullet \cite{pybullet} and MuJoCo \cite{todorov2012mujoco}. However, given that these are physics engines rather than robotics frameworks, it can often be cumbersome to build rich environments and integrate standard robotics tooling such as inverse and forward kinematics, user interfaces, motion libraries, and path planners. Given the scale of RLBench, we needed a tool for designing tasks as easily as possible.

The task building tool is the interface for users who wish to create new tasks to be added to the RLBench task repository. Each task has 2 associated files: a V-REP model file (\textit{.ttm}), which holds all of the scene information and demo waypoints, and a python (\textit{.py}) file, which is responsible for wiring the scene objects to the RLBench backend, applying variations, defining success criteria, and adding other more complex task behaviours. Figure \ref{alg:task_builder_example} shows an example of how simple many tasks files can be.

\begin{figure*}
  \centering
  \begin{subfigure}[b]{0.99\textwidth}
    \includegraphics[width=0.99\linewidth]{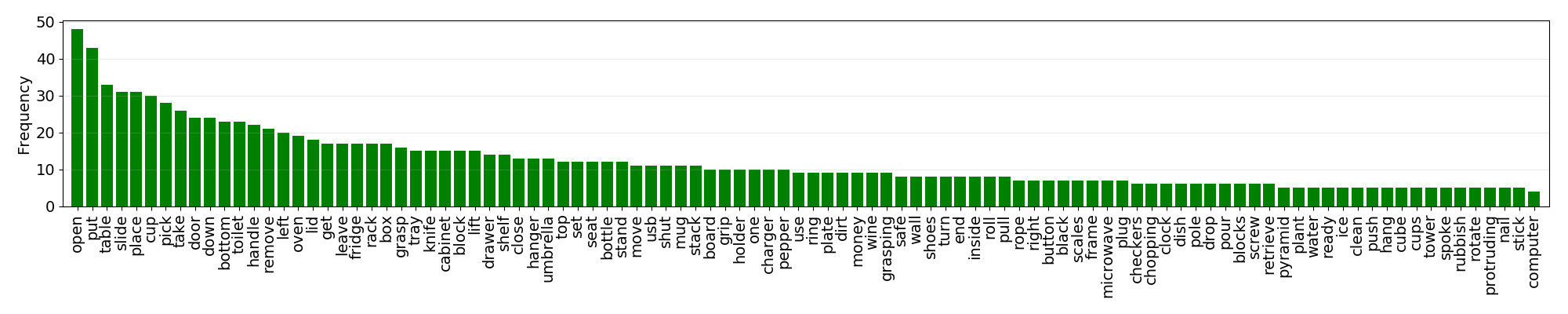}        
  \end{subfigure}
  \begin{subfigure}[b]{0.99\textwidth}
    \includegraphics[width=0.99\linewidth]{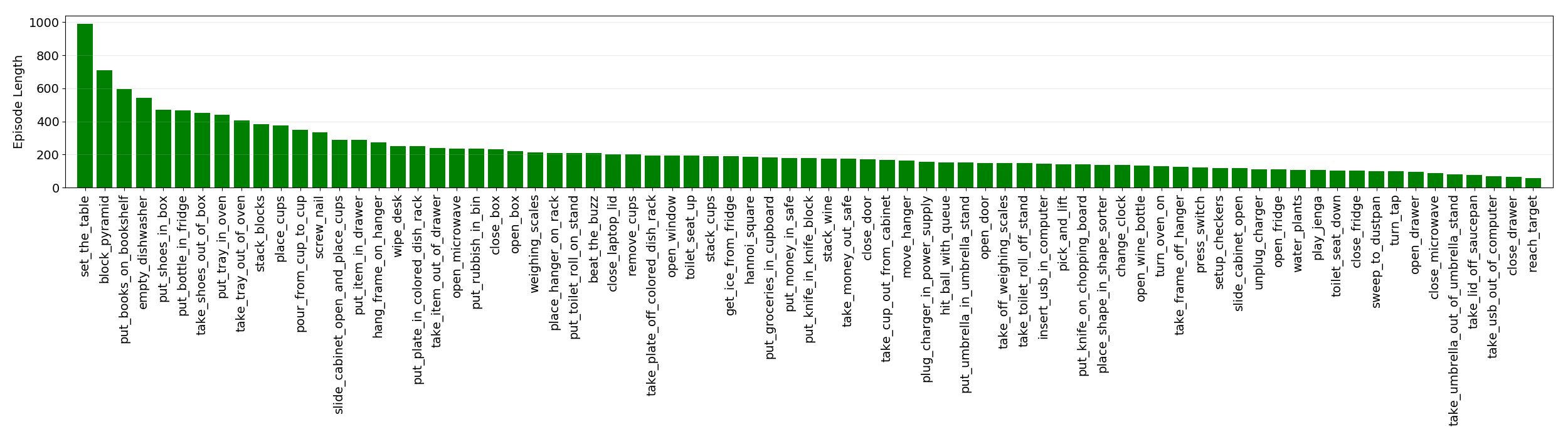}      
  \end{subfigure}
  \caption{\textbf{Top} shows the frequency of words in the variation descriptions with function words removed, leaving only content words. \textbf{Bottom} shows the average length of 5 demonstrations from a sample of 75 tasks (taken from the first variation). The tasks lengths vary from $100$ to $1000$ timesteps. Longer tasks usually involve many composed sets of actions, for example, the `empty\_dishwasher' task involves opening the washer door, sliding out the tray, grasping a plate, and then lifting the plate out of the tray. These long-horizon tasks can facilitate interesting research in reinforcement learning in robotic tasks.}
  \label{fig:word_plot_and_demo_lens_plot}
\end{figure*}

In order to use the task creator, users must understand how tasks are initialised and placed in the scene. When a user asks for a new task from RLBench, the task is initialised by calling $init\_task()$, and is only called once.  Following that,  $init\_variation(int\text{ }i)$ is called at the beginning of each variation, and gets passed the variation number, which should be less than or equal to the number of variations for that task (which can be obtained by calling $variation\_count()$). This function returns a list of strings which provide descriptions that could be associated with this variation of the task; an analysis of the frequency of words in these descriptions can be seen in top of Figure \ref{fig:word_plot_and_demo_lens_plot}. Finally, $init\_episode()$ is called each time a new episode (of the same variation) is requested.

\begin{figure}[]
\begin{lstlisting}[language=Python]
from rlbench.environment import Environment
from rlbench.action_modes import ActionMode
from rlbench.tasks import ReachTarget

DATASET = 'path/to/demo/dataset'

env = Environment(
  DATASET, ActionMode.ABS_JOINT_VELOCITY)
env.launch()

task = env.sample_task()
demos = task.get_demos(2)

agent = Agent()
agent.ingest(demos)

training_steps = 100
episode_length = 100
obs = None
for i in range(training_steps):
  if i % episode_length == 0:
    descriptions, obs = task.reset()
  action = agent.act(obs)
  obs, reward, terminate = task.step(action)
env.shutdown()
\end{lstlisting}
\caption {Example usage of the RLBench Environment for training a reinforcement learning agent. When using demonstrations, users can either point to a set of saved demonstrations (as shown here), or alternatively generate demonstrations on the fly.}
\label{alg:api_example}
\end{figure}

\begin{figure}[]
\begin{lstlisting}[language=Python]
from rlbench.backend.task import Task
from rlbench.backend.conditions import DetectedCondition, GraspedCondition
from pyrep.objects.shape import Shape
from pyrep.objects.proximity_sensor import ProximitySensor

class TakeLidOffSaucepan(Task):

  def init_task(self):
    lid = Shape('saucepan_lid')
    success_detector = ProximitySensor('success')
    self.register_graspable_objects([lid])
    cond_set = [
      GraspedCondition(self.robot.gripper, lid),
      DetectedCondition(lid, success_detector)
    ]
    self.register_success_conditions([cond_set])

  def init_episode(self, index):
    return ['take lid off the saucepan']

  def variation_count(self):
    return 1
        
\end{lstlisting}
\caption {An example of a task python file. When using the task building tool, users are able to simultaneously edit the V-REP scene whilst also changing the various behaviour of a task. In this example, the task is to take a lid off of a saucepan. By interfacing with the scene using PyRep, we register that the episode should terminate and be considered a success only if the saucepan lid is detected by a proximity sensor and that the lid is being held. The backend handles the randomisation of the position of the task at the beginning of each episode.}
\label{alg:task_builder_example}
\end{figure}

Once a task has been created, we provide a task validation tool, that attempts to collect a number of demonstrations of the designed task in order to ensure that the path planning aspect of the task only fails a small number of times. Once the validator passes, the user will be free to perform a GitHub pull request in order to contribute to the growing task repository.

\section{The RLBench Few-Shot Challenge ($v\text{ }1.0$)}
\label{sec:fewshot_challenge}

A big gap in the literature today is a means to evaluate and compare few-shot learning methods for robotics. We place particular emphasis on the few-shot regime, because much like humans, robots should have the ability to leverage knowledge from previously learned tasks in order to learn new ones quickly in new and unfamiliar environments. Despite this, most approaches in manipulation have focused on learning a single task, with a limited notion of generalisation, and no way of leveraging the knowledge to learn other tasks more efficiently.

The few pieces of work that perform few-shot learning in robotics \cite{finn2017one,  james2018task, yu2018one} focused on a very narrow definition of task and often treat a variation of the same task as another task; for example, placing a peach into a red bowl would be considered a different task to placing an apple into a green bowl. In order to develop truly general algorithms, we feel that it is important to have a diverse range of tasks to train and test on. To that end, we propose the following challenge:

Given N unseen tasks, provide the system with K different demonstrations of each of the N tasks, and then evaluate the systems ability to perform these tasks in new configurations. Specifically, we suggest the following procedure:

\begin{itemize}
    \item Of the \numtasks{} unique tasks, $10\%$ of the tasks have been selected for the test set (meta-test) which span a range of difficulties, while the rest are chosen for training (meta-train). These train-test splits will be made available on the benchmark's webpage.
    \item The training tasks can be used in any way desired by the user. RLBench supplies a large number of pre-generated demos for each task that can be downloaded, although there is also the option to generate demos on the fly (or for users to create their own).
    \item During test time, the system is given K demonstrations of the unseen task (K-shot), and then success should be reported on new episodes of that same task. The only information available to the system should be the number of demos N and their corresponding observations. There must be no prior knowledge of the unseen tasks given to the system that are not included in the training tasks. Users report 1-shot, 5-shot, and 20-shot results for their method. 
\end{itemize}

We purposefully call this challenge $v\text{ }1.0$ as we expect the number of tasks to grow considerably over the years; as this happens, we will create newer versions that span a broader range of tasks; therefore, we hope this versioning will ensure results remain meaningful and reproducible as the benchmark grows. State-of-the-art few-shot learning methods such as recurrent methods \cite{santoro2016meta, duan2016rl, mishra2017simple}, metric learning methods \cite{vinyals2016matching, snell2017prototypical}, and gradient based methods \cite{finn2017model, rusu2018meta} have not been tested on such a grand scale, and we look forward to seeing how they perform on this benchmark.

\section{Other Applications \& Challenges}

Further to the few-shot learning challenge highlighted in Section \ref{sec:fewshot_challenge}, we briefly overview other areas of research that could benefit from RLBench.

\paragraph{\textbf{Reinforcement Learning}} 

There is a large body of work in continuous control reinforcement learning that evaluate their algorithms on benchmarks such as OpenAI Gym \cite{brockman2016openai} or DeepMind Control Suite \cite{tassa2018deepmind}. Unlike these benchmarks, RLBench has been tailored for visually-guided manipulation, which makes this an ideal platform for evaluating current and future reinforcement learning algorithms on real-world based tasks. Moreover, given the large number of demonstrations provided, it opens up the space to accelerate and facilitate research in bootstrapping reinforcement learning policies with demonstrations in order to reduce sample complexity. In addition, with the provided eye-in-hand camera observations, we open research in partial observability or incremental estimation for continuous control tasks.

\paragraph{\textbf{Imitation Learning}} 

Almost all imitation learning work design their own tasks for evaluating their method, making reproducibility difficult. A set number of demonstrations are shipped with RLBench, but there is also the option in the framework to generate demonstrations on-the-fly, meaning that you cam generate an infinite amount for your imitation learning algorithm. 

\paragraph{\textbf{Sim-to-Real Transfer}} 

Recently there has been a large amount of work in learning control policies in simulation and then transferring these to the real world \cite{james2017transferring, peng2018sim, matas2018sim, hwangbo2019learning, bousmalis2018using, james2019sim}.
The simulated Franka Panda within RLBench can be easily swapped out, with one line of code, for another arm that researchers may have in their lab; this means that sim-to-real methods could be compared more easily on a standard set of tasks. Moreover, given the task-building tool and demonstration generation that RLbench has to offer, new tasks can easily be designed to demonstrate particular features in novel sim-to-real methods. 

\paragraph{\textbf{Multi-task Learning}} 

In contrast to few-shot learning, multi-task learning concerns itself with learning several tasks simultaneously without particularly being expected to generalise to radically different tasks at test time. In this setup, all tasks from both meta-training and meta-testing can be used during training, and then during testing, the system must be able to generalise to unseen examples of those tasks. Given the difficulty of the challenge laid out in Section \ref{sec:fewshot_challenge}, tackling the multi-task problem could provide valuable insights to increasing performance in the few-shot domain. 

\paragraph{\textbf{SLAM}} 

Simultaneous Localisation and Mapping (SLAM) is concerned with constructing a map of an unknown environment while simultaneously keeping track of an agent's location within it. Traditionally SLAM has been limited to navigation, virtual reality and augmented reality domains; but ultimately we can envision SLAM systems playing a key role in robots interacting with the world, i.e. a focus on more task-based SLAM. However, if we would like a manipulation system to make use of a SLAM map, it is not currently clear what the best way to represent this map is: whether it be sparse \cite{mur2017orb, forster2014svo}, dense \cite{newcombe2011dtam, Newcombe:etal:ISMAR2011}, or semi-dense \cite{engel2014lsd}. Moreover, it is not clear what level accuracy the map would need in order to achieve a desired task. RLBench could facilitate research in unifying SLAM and manipulation more tightly.

\section{Summary and Future Work}

We have presented RLBench, an attempt to accelerate research in robotic manipulation that can be used in a broad range of robotic related research. We have posed the few-shot learning challenge for manipulation, and have highlighted a number of research areas that could benefit from this large scale benchmark and learning environment.

Given the scale of this project, we envision that there may be teething problems as people begin using the platform, and so we aim to maintain and continuously improve the benchmark during launch. Further to that, we hope, along with the help of the community, to continuously expand the tasks available for both training and evaluation. We hope RLBench will become a key resource for a broad range of robot manipulation related research, and look forward to seeing what the community achieves with this diverse range of tasks.

\section*{ACKNOWLEDGMENTS}
We thank Juxi Leitner, Ankur Handa and Eugene Valassakis for insightful feedback on an early draft of this paper. Research presented here has been supported by Dyson Technology Ltd.


\bibliographystyle{IEEEtran}
\bibliography{ref}

\end{document}